%% file: main.tex
\documentclass[10pt,twocolumn,letterpaper]{article}

\usepackage{cvpr}
\usepackage{times}
\usepackage{epsfig}
\usepackage{graphicx}
\usepackage{amsmath}
\usepackage{amssymb}
\usepackage{multirow}
\usepackage{bm}
\usepackage[OT1]{fontenc} 
\usepackage[ruled]{algorithm2e}
\usepackage{caption}
\usepackage{graphicx}
\usepackage{subfigure}

\usepackage[pagebackref=true,breaklinks=true,letterpaper=true,colorlinks,bookmarks=false]{hyperref}

\cvprfinalcopy % *** Uncomment this line for the final submission

\begin{document}

%%%%%%%%% TITLE
\title{Fast and Robust Multi-Person 3D Pose Estimation from Multiple Views}

\author{Junting Dong  \\
	Zhejiang University\\
	%\\
	%Hangzhou, China, 310058\\
	{\tt\small jtdong@zju.edu.cn}
	\and
	Wen Jiang\\
	Zhejiang University\\
	%\\
	%Hangzhou, China, 310058\\
	{\tt\small wenjiang@zju.edu.cn}
	\and
	Qixing Huang\\
	University of Texas at Austin\\
	%Department of Computer Science\\
	%Austin, TX 78712, USA \\
	{\tt\small huangqx@cs.utexas.edu}
	\and
	Hujun Bao\\
	Zhejiang University\\
	%\\
	%Hangzhou, China, 310058\\
	{\tt\small bao@cad.zju.edu.cn}
	\and
	Xiaowei Zhou\\
	Zhejiang University\\
	%Department of Computer Science\\
	%The University of Texas at Austin\\
	%Austin, TX 78712, USA \\
	{\tt\small xzhou@cad.zju.edu.cn}
}

\maketitle
%\thispagestyle{empty}

%%%%%%%%% ABSTRACT
\begin{abstract}
\input{00_abstract.tex}

\end{abstract}

%%%%%%%%% BODY TEXT
\section{Introduction}
\input{01_intro}

%------------------------------------------------------------------------
\section{Related work}

\input{02_related_work}

%------------------------------------------------------------------------
\section{Technical approach}

\input{03_technical_approach}

\section{Empirical evaluation}

\input{04_empirical_evaluation}

%------------------------------------------------------------------------
\section{Summary}

\input{05_summary}

\vfill

{\small
\bibliographystyle{ieee}
\bibliography{egbib}
}

\end{document}

%% file: 00_abstract.tex
%!TEX root = ./egpaper_for_review.tex

This paper addresses the problem of 3D pose estimation for multiple people in a few calibrated camera views. The main challenge of this problem is to find the cross-view correspondences among noisy and incomplete 2D pose predictions. Most previous methods address this challenge by directly reasoning in 3D using a pictorial structure model, which is inefficient due to the huge state space. We propose a fast and robust approach to solve this problem. Our key idea is to use a multi-way matching algorithm to cluster the detected 2D poses in all views. Each resulting cluster encodes 2D poses of the same person across different views and consistent correspondences across the keypoints, from which the 3D pose of each person can be effectively inferred. The proposed convex optimization based multi-way matching algorithm is efficient and robust against missing and false detections, without knowing the number of people in the scene. Moreover, we propose to combine geometric and appearance cues for cross-view matching. The proposed approach achieves significant performance gains from the state-of-the-art ($96.3\%$ vs. $90.6\%$ and $96.9\%$ vs. $88\%$ on the Campus and Shelf datasets, respectively), while being efficient for real-time applications.   

%% file: 01_intro.tex
%!TEX root = ./egpaper_for_review.tex

\begin{figure}[t]
	\centering
	\includegraphics[width=1\linewidth,trim={5.5cm 0cm 7.5cm 0cm},clip]{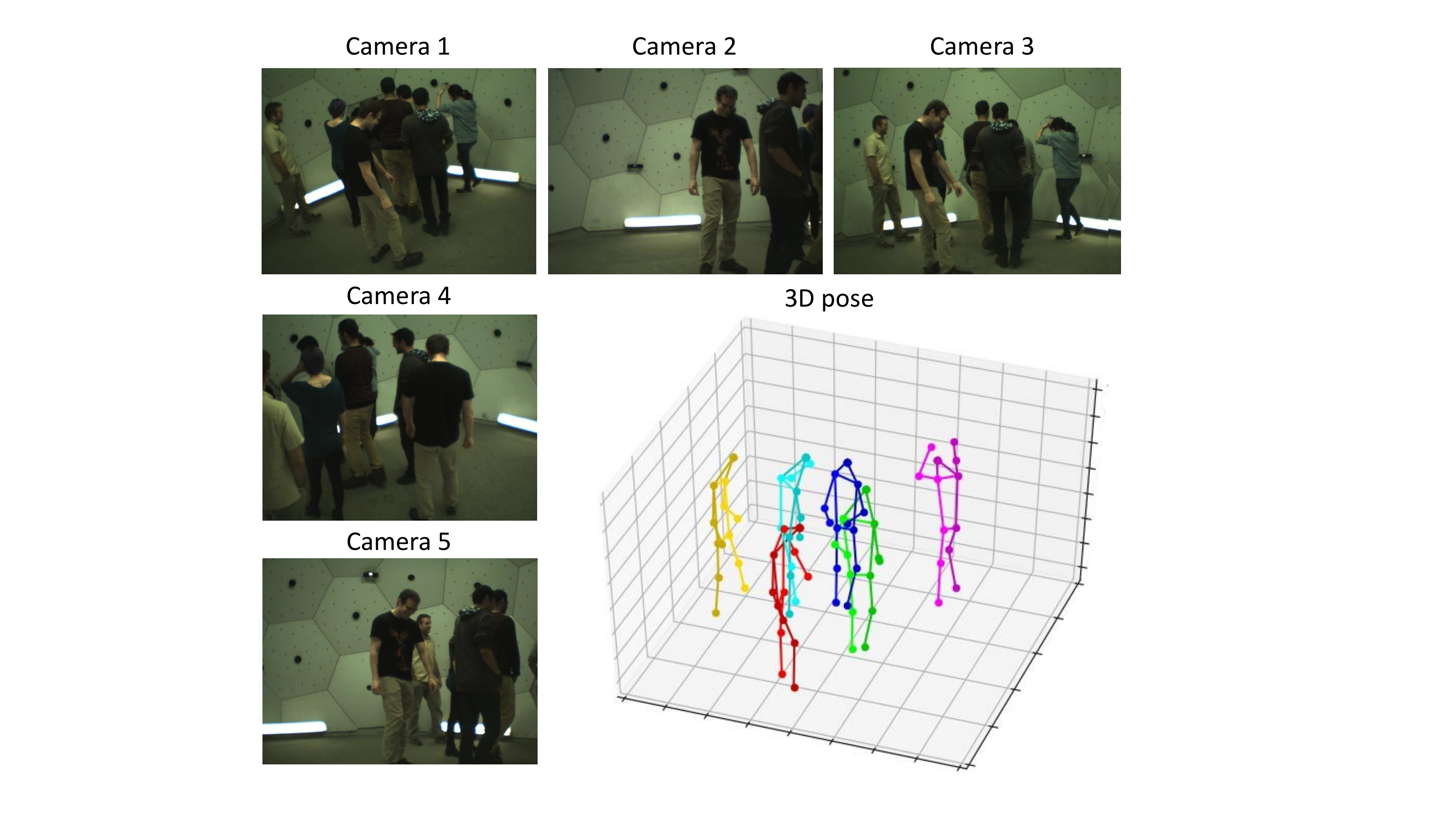}
	\caption{
	This work proposes a novel approach for fast and robust recovery of 3D poses of multiple people from a few camera views. The main challenge is to establish consistent correspondences of 2D observations among multiple views, e.g., 2D human-body keypoints in images, which may be noisy and incomplete.
	}
	\label{fig:result}
\end{figure}

Recovering 3D human pose and motion from videos has been a long-standing problem in computer vision, which has a variety of applications such as human-computer interaction, video surveillance and sports broadcasting. In particular, this paper focuses on the setting where there are multiple people in a scene, and the observations come from a few calibrated cameras (See Figure \ref{fig:result}). While remarkable advances have been made in multi-view reconstruction of a human body, there are fewer works that address a more challenging setting where multiple people interact with each other in crowded scenes, in which there are significant occlusions.

Existing methods typically solve this problem in two stages. The first stage detects human-body keypoints or parts in separate 2D views, which are aggregated in the second stage to reconstruct 3D poses. Given the fact that deep-learning based 2D keypoint detection techniques have achieved remarkable performance~\cite{cao2016realtime,newell2016stacked}, the remaining challenge is to find the cross-view correspondences between detected keypoints as well as which person they belong to. Most previous methods \cite{belagiannis20143d,belagiannis20163d,joo2017panoptic,ershadi2018multiple} employ a 3D pictorial structure (3DPS) model that implicitly solves the correspondence problem by reasoning about all hypotheses in 3D that are geometrically compatible with 2D detections. However, this 3DPS-based approach is computational expensive due to the huge state space. In addition, it is not robust particularly when the number of cameras is small, as it only uses multi-view geometry to link the 2D detections across views, or in other words, the appearance cues are ignored.

\begin{figure*}
	% minipage mit (Blind-)Text
	\centering
	\includegraphics[width=1\linewidth,trim={0.0cm 5cm 1cm 1cm},clip]{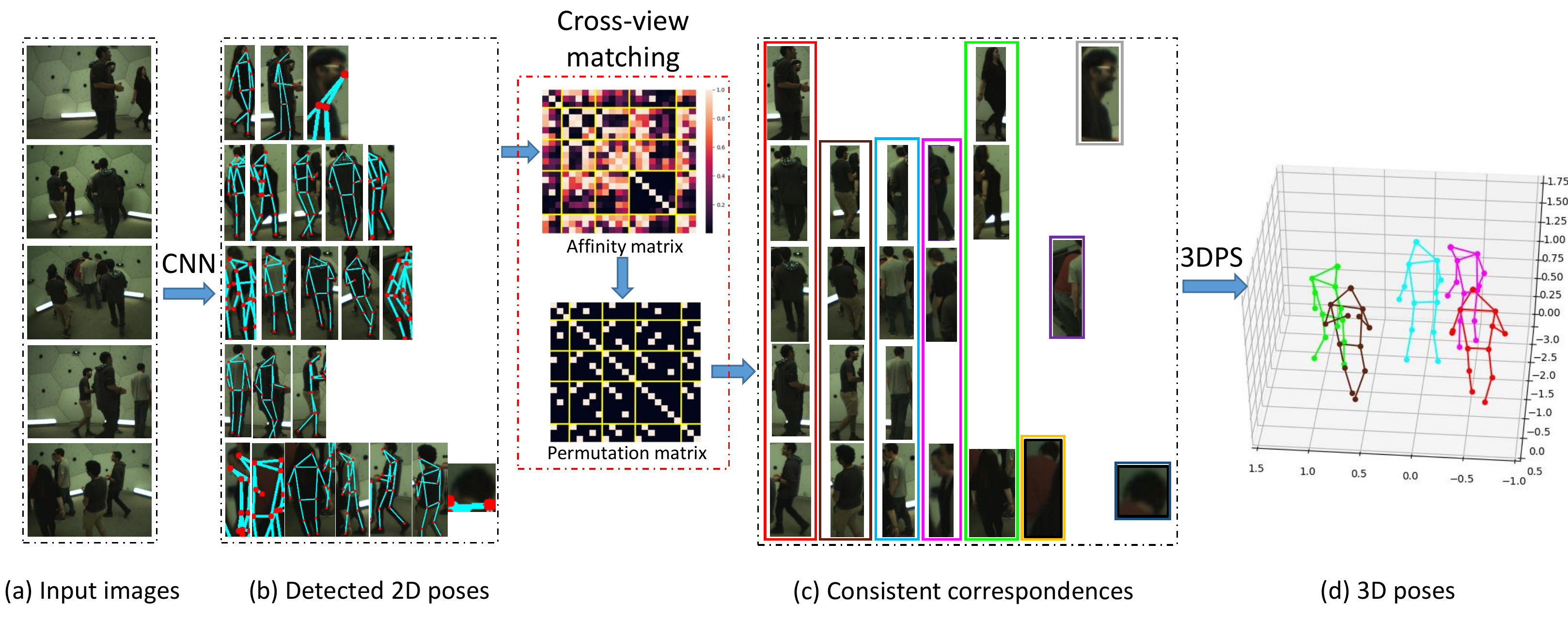}
	\caption{
		Overview of the proposed approach. Given images from a few calibrated cameras (a), an off-the-shelf human pose detector is used to produce 2D bounding boxes and associated 2D poses in each view, which may be inaccurate and incomplete (b). Then, the detected bounding boxes are clustered by a novel multi-view matching algorithm. Each resulting cluster includes the bounding boxes of the same person in different views (c). The isolated bounding boxes that have no matches in other views are regarded as false detections and discarded. Finally, the 3D pose of each person is reconstructed from the corresponding bounding boxes and associated 2D poses (d).
	}
	
	\label{fig:pipeline}
\end{figure*}

In this paper, we propose a novel approach for multi-person 3D pose estimation. The proposed approach solves the correspondence problem at the body level by matching detected 2D poses among multiple views, producing clusters of 2D poses where each cluster includes 2D poses of the same person in different views. Then, the 3D pose can be inferred for each person separately from matched 2D poses, which is much faster than joint inference of multiple poses thanks to the reduced state space.

However, matching 2D poses across multiple views is challenging. A typical approach is to use the epipolar constraint to verify if two 2D poses are projections of the same 3D pose for each pair of views \cite{kadkhodamohammadi2018generalizable}. But this approach may fail for the following reasons. First, the detected 2D poses are often inaccurate due to heavy occlusion and truncation, as shown in Figure~\ref{fig:pipeline}(b), which makes geometric verification difficult. Second, matching each pair of views separately may produce inconsistent correspondences which violate the cycle consistency constraint, that is, two corresponding poses in two views may be matched to different people in another view. Such inconsistency leads to incorrect multi-view reconstructions. Finally, as shown in Figure~\ref{fig:pipeline}, different sets of people appear in different views and the total number of people is unknown, which brings additional difficulties to the matching problem.

We propose a multi-way matching algorithm to address the aforementioned challenges. Our key ideas are: (i) combing the geometric consistency between 2D poses with the appearance similarity among their associated image patches to reduce matching ambiguities, and (ii) solving the matching problem for all views simultaneously with a cycle-consistency constraint to leverage multi-way information and produce globally consistent correspondences. The matching problem is formulated as a convex optimization problem and an efficient algorithm is developed to solve the induced optimization problem.

In summary, the main contributions of this work are:
\begin{itemize}
\item We propose a novel approach for fast and robust multi-person 3D pose estimation. We demonstrate that, instead of jointly inferring multiple 3D poses using a 3DPS model in a huge state space, we can greatly reduce the state space and consequently improve both efficiency and robustness of 3D pose estimation by grouping the detected 2D poses that belong to the same person in all views.
\item We propose a multi-way matching algorithm to find the cycle-consistent correspondences of detected 2D poses across multiple views. The proposed matching algorithm is able to prune false detections and deal with partial overlaps between views, without  knowing the true number of people in the scene.
\item We propose to combine geometric and appearance cues to match the detected 2D poses across views. We show that the appearance information, which is mostly ignored by previous methods, is important to link the 2D detections across views.
\item The proposed approach outperforms the state-of-the-art methods by a large margin without using any training data from the evaluated datasets. The code will be available upon publication at \url{https://zju-3dv.github.io/mvpose/}.

\end{itemize}

%% file: 02_related_work.tex
%!TEX root = ./egpaper_for_review.tex

\paragraph{\bf Multi-view 3D human pose:} Markerless motion capture has been investigated in computer vision for a decade. Early works on this problem aim to track the 3D skeleton or geometric model of human body through a multi-view sequence \cite{taylor2010dynamical,yao2011learning,elhayek2015efficient}. These tracking-based methods require initialization in the first frame and are prone to local optima and tracking failures. Therefore, more recent works are generally based on a bottom-up scheme where the 3D pose is reconstructed from 2D features detected from images \cite{sigal2012,burenius20133d,pavlakos2017harvesting}. Recent work \cite{joo2018total} shows remarkable results by combing statistical body models with deep learning based 2D detectors.

In this work, we focus on the multi-person 3D pose estimation. Most previous works are based on 3DPS models in which nodes represent 3D locations of body joints and edges encode pairwise relations between them \cite{belagiannis20143d,joo2015panoptic,belagiannis20163d,joo2017panoptic,ershadi2018multiple}. The state space for each joint is often a 3D grid representing a discretized 3D space. The likelihood of a joint being at some location is given by a joint detector applied to all 2D views and the pairwise potentials between joints are given by skeletal constraints \cite{belagiannis20143d,belagiannis20163d} or body parts detected in 2D views \cite{joo2017panoptic,ershadi2018multiple}. Then, the 3D poses of multiple people are jointly inferred by maximum a posteriori estimation. 

As all body joints for all people are considered simultaneously, the entire state space is huge, resulting in heavy computation in inference. Another limitation of this approach is that it only uses multi-view geometry to link 2D evidences, which is sensitive to the setup of cameras. As a result, the performance of this approach degrades significantly when the number of views decreases \cite{joo2017panoptic}. Recent work \cite{kadkhodamohammadi2018generalizable} proposes to match 2D poses between views and then reconstruct 3D poses from the 2D poses belonging to the same person. But it only utilizes epipolar geometry to match 2D poses for each pair of views and ignores the cycle consistency constraint among multiple views, which may result in inconsistent correspondences.  

\paragraph{\bf Single-view pose estimation:} There is a large body of literature on human pose estimation from single images. Single-person pose estimation \cite{toshev2014deeppose,pfister2015flowing,wei2016convolutional,newell2016stacked,huang2017coarse} localizes 2D body keypoints of a person in a cropped image. There are two categories of multi-person pose estimation methods: top-down methods \cite{chen2017cascaded,huang2017coarse,he2018mask,fang2017rmpe} that first detect people in the image and then apply single-person pose estimation to the cropped image of each person, and bottom-up methods \cite{kocabas2018multiposenet,newell2017associative,cao2016realtime,pishchulin2016deepcut,insafutdinov2016deepercut} that first detect all keypoints and then group them into different people. In general, the top-down methods are more accurate, while the bottom-up methods are relatively faster. In this work, We adopt the Cascaded Pyramid Network \cite{chen2017cascaded}, a state-of-the-art approach for multi-person pose detection, as an initial step in our pipeline. 

The advances in learning-based methods also make it possible to recover 3D human pose from a single RGB image, either lifting the detected 2D poses into 3D \cite{moreno20173d,zhou2016sparseness,chen20173d,martinez2017simple} or directly regressing 3D poses \cite{tome2017lifting,sun2017compositional,tekin2017learning,zhou2017towards,pavlakos2018ordinal} and even 3D body shapes from RGB \cite{bogo2016keep,kanazawa2018end,pavlakos2018learning}. But the reconstruction accuracy of these methods is not comparable with the multi-view results due to the inherit reconstruction ambiguity when only a single view is available.  

\paragraph{\bf Person re-ID and multi-image matching:} Person re-ID aims to identify the same person in different images \cite{zhong2018camera}, which is used as a component in our approach. Multi-image matching is to find feature correspondences among a collection of images \cite{huang2013consistent,zhou2015multi}. We make use of the recent results on cycle consistency \cite{huang2013consistent} to solve the correspondence problem in multi-view pose estimation.

%% file: 03_technical_approach.tex
%!TEX root = ./egpaper_for_review.tex

Figure \ref{fig:pipeline} presents an overview of our approach. First, an off-the-shelf 2D human pose detector is adopted to produce bounding boxes and 2D keypoint locations of people in each view (Section \ref{sec:generic 2D}). Given the noisy 2D detections, a multi-way matching algorithm is proposed to establish the correspondences of the detected bounding boxes across views and get rid of the false detections (Section \ref{sec:multi-view people matching}). Finally, the 3DPS model is used to reconstruct the 3D pose for each person from the corresponding 2D bounding boxes and keypoints (Section \ref{sec:multi-view}).

%-------------------------------------------------------------------------
\subsection{2D human pose detection}\label{sec:generic 2D}

We adopt the recently-proposed Cascaded Pyramid Network \cite{chen2017cascaded} trained on the MSCOCO \cite{linmicrosoft} dataset for 2D pose detection in images. The Cascaded Pyramid Network consists of two stages: the GlobalNet estimates human poses roughly whereas the RefineNet gives optimal human poses. Despite its state-of-the-art performance on benchmarks, the detections may be quite noisy as shown in Figure \ref{fig:pipeline}(b).

\subsection{Multi-view correspondences}\label{sec:multi-view people matching}
\iffalse
\begin{figure}[t]
	\centering
	\includegraphics[width=0.99\linewidth,trim={5cm 7cm 4cm 3cm},clip]{Figures/resource/matchSVT2.pdf}
	\caption{
		The procedure of our consistent multi-view correspondences. Given images from $N$ views as input, we first use 2D human pose detection to acquire bounding boxes of candidate people, which are very noisy. With the proposed SVT matching algorithm,  we finally obtains globally consistent correspondences. The bounding boxes inside the dotted box is considered to belong to the same person.
	}
	\label{fig:svt}
\end{figure}
\fi
Before reconstructing the 3D poses, the detected 2D poses should be matched across views, i.e., we need to find in all views the 2D bounding boxes belonging to the same person. However, this is a challenging task as we discussed in the introduction.

To solve this problem, we need 1) a proper metric to measure the likelihood that two 2D bounding boxes belong to the same person (a.k.a. affinity), and 2) a matching algorithm to establish the correspondences of bounding boxes across multiple views. In particular, the matching algorithm should not place any assumption about the true number of people in the scene. Moreover, the output of the matching algorithm should be cycle-consistent, i.e. any two corresponding bounding boxes in two images should correspond to the same bounding box in another image. 

\paragraph{Problem statement:}
Before introducing our approach in details, we first briefly describe some notations. Suppose there are $V$ cameras in the scene and $p_i$ detected bounding boxes in view $i$. For a pair of views $(i,j)$, the affinity scores can be calculated between the two sets of bounding boxes in view $i$ and view $j$. We use $\bm{A}_{ij}\in \mathbb R^{p_i \times p_j}$ to denote the affinity matrix, whose elements represent the affinity scores. The correspondences to be estimated between the two sets of bounding boxes are represented by a partial permutation matrix  $\bm{P}_{ij} \in \{0,1\}^{p_i \times p_j}$, which satisfies the doubly stochastic constraints:
\begin{equation}\label{eq:doubly stochastic constraints}
\bm{0} \leq \bm{P}_{ij}\bm{1} \leq \bm{1}, \bm{0} \leq \bm{P}_{ij}^T\bm{1} \leq \bm{1}.
\end{equation} 
The problem is to take $\{\bm{A}_{ij}|\forall i,j\}$ as input and output the optimal $\{\bm{P}_{ij}|\forall i,j\}$
that maximizes the corresponding affinities and is also cycle-consistent across multiple views. 

\paragraph{Affinity matrix:} We propose to combine the appearance similarity and the geometric compatibility to calculate the affinity scores between bounding boxes. 

First, we adopt a pre-trained person re-identification (re-ID) network to obtain a descriptor for a bounding box. The re-ID network trained on massive re-ID datasets is expected to be able to extract discriminative appearance features that are relatively invariant to illumination and viewpoint changes. Specifically, we feed the cropped image of each bounding box through the publicly available re-ID model proposed in \cite{zhong2018camera} and extract the feature vector from the ``pool5'' layer as the descriptor for each bounding box. Then, we compute the Euclidean distance between the descriptors of a bounding box pair and map the distances to values in $(0, 1)$ using the sigmoid function as the appearance affinity score of this bounding box pair. 

Besides appearances, another important cue to associate two bounding boxes is that their associated 2D poses should be geometrically consistent. Specifically, the corresponding 2D joint locations should satisfy the epipolar constraint, i.e. a joint in the first view should lie on the epipolar line associated with its correspondence in the second view. Suppose $\bm{x} \in \mathbb{R}^{N \times 2}$ denotes a 2D pose composed of $N$ joints. Then, the geometric consistency between $\bm{x}_i$ and $\bm{x}_j$ from two views can be measured by the following distance:   
\begin{equation}
\begin{split}
D_g(\bm{x}_i,\bm{x}_j) = \frac{1}{2N} \sum_{n=1}^{N} d_g(\bm{x}^n_i,\bm{L}_{ij}(\bm{x}^n_j))+d_g(\bm{x}^n_j,\bm{L}_{ji}(\bm{x}^n_i)), \nonumber
\end{split}
\end{equation}
where $\bm{x}^n_i$ denotes the 2D location of the $n$-th joint of pose $i$, $\bm{L}_{ij}(\bm{x}_j^n)$ the epipolar line associated with $\bm{x}_j^n$ from the other view, and $d_g(\cdot,\bm{l})$ the point-to-line distance for $l$. The distances $D_g$ are also mapped to values in $(0, 1)$ using the sigmoid function as the final geometric affinity scores.  

Based on the fact that a pair of correctly detected and matched 2D poses must satisfy the geometric constraint ($D_g$ is small), we combine the two affinity matrices as follows:
\begin{equation}
\bm{A}_{ij}(\cdot) = \left\{
\begin{aligned}
\sqrt{\bm{A}_{ij}^a(\cdot) \times \bm{A}_{ij}^g(\cdot)}, ~~~~~~ &\mbox{if}~~ D_g \leq th, \\
0, \qquad \qquad ~~ &\mbox{otherwise},
\end{aligned}
\right.
\end{equation} 
where $\bm{A}_{ij}(\cdot)$, $\bm{A}_{ij}^a(\cdot)$, and $\bm{A}_{ij}^g(\cdot) \in [0,1]$ denote values of the fused affinity matrix, appearance affinity matrix, and geometry affinity matrix of view pair $(i,j)$, respectively. $th$ denotes a threshold. Experimental results demonstrate that this simple combination of appearance and geometry is superior to merely using one of them. 

\begin{figure}[t]
	\centering
	\includegraphics[width=0.99\linewidth,trim={6.5cm 3.5cm 8cm 1cm},clip]{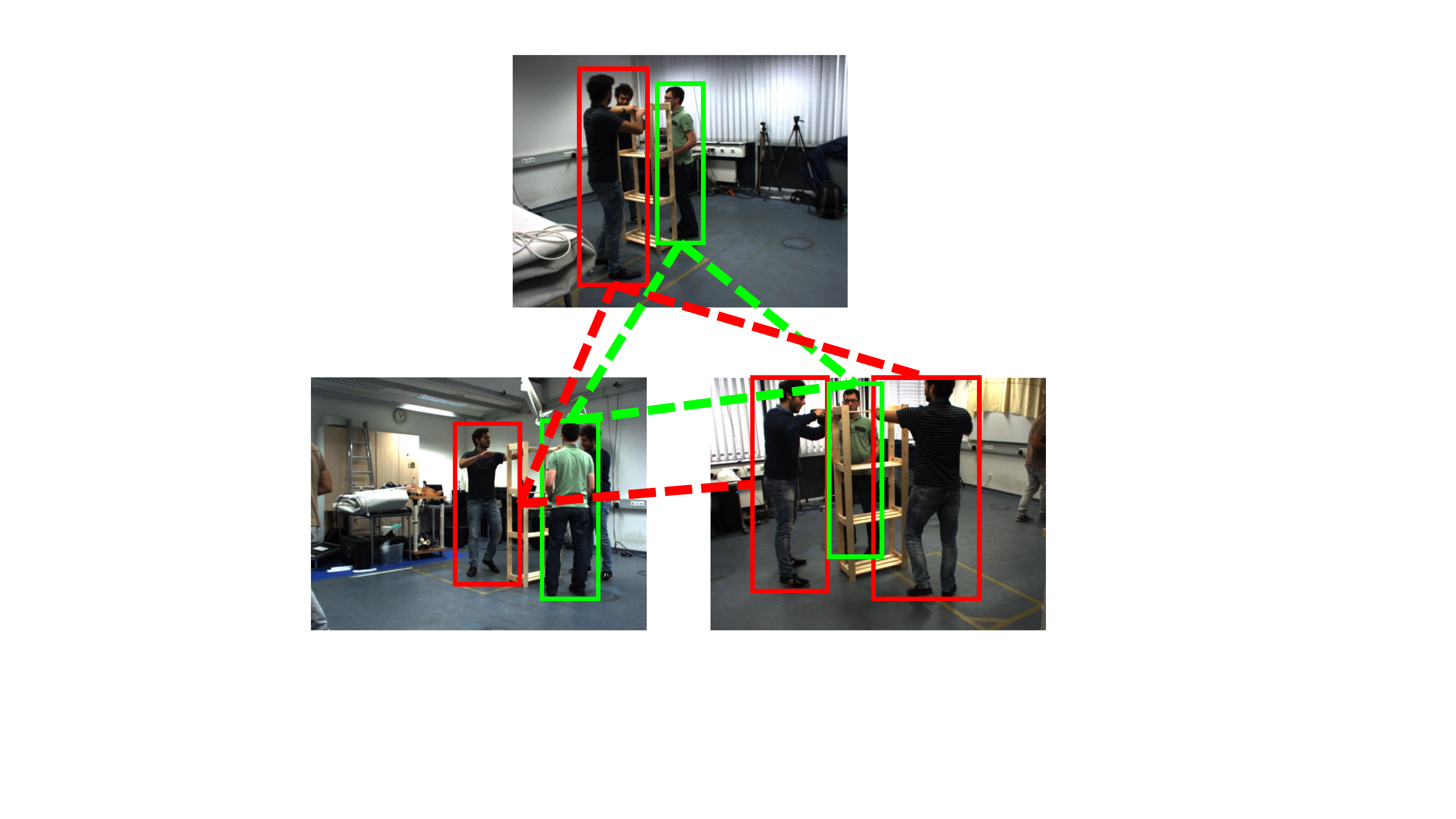}
	\caption{
		An illustration of cycle consistency. The green lines denote a set of consistent correspondences and the red lines show a set of inconsistent correspondences.  
	}
	\label{fig:cycle}
\end{figure}

\paragraph{Multi-way matching with cycle consistency:} If there are only two views to match, one can simply maximize $\langle \bm{P}_{ij},\bm{A}_{ij} \rangle$ and find the optimal matching by the Hungarian algorithm. But when there are multiple views, solving the matching problem separately for each pair of views ignores the cycle-consistency constraint and may lead to inconsistent results. Figure~\ref{fig:cycle} shows an example, where the correspondences in red are inconsistent and the ones in green are cycle-consistent as they form a closed cycle. 

We make use of the results in \cite{huang2013consistent} to solve this problem. Suppose the correspondences among all $m = \sum_{i=1}^V p_i$ detected bounding boxes in all views are denoted by $\bm{P} \in \{0,1\}^{m \times m}$:
\begin{equation}\label{eq:big_p}
\bm{P} = 
\left(
\begin{matrix}
\bm{P}_{11} & \bm{P}_{12} & \cdots & \bm{P}_{1n} \\
\bm{P}_{21} & \bm{P}_{22} & \cdots & \bm{P}_{2n} \\
\vdots & \vdots & \ddots & \vdots \\
\bm{P}_{n1} & \cdots & \cdots & \bm{P}_{nn}
\end{matrix}
\right), 
\end{equation}
where $\bm{P}_{ii}$ should be identity. 
Then, it can be shown that the cycle consistency constraint is satisfied if and only if
\begin{equation}\label{eq:low rank}
\mbox{rank}(\bm{P}) \leq s, \ \bm{P} \succeq 0,
\end{equation}
where $s$ is the underlying number of people in the scene. The intuition is that, if the correspondences are cycle-consistent, $\bm{P}$ can be factorized as $\bm{Y}\bm{Y}^T$ where $\bm{Y}\in\mathbb{R}^{m \times s}$ denotes the correspondences between all 2D bounding boxes and 3D people. 

As $s$ is unknown in advance, we propose to minimize the following objective function to estimate the low-rank and positive semidefinite matrix $\bm{P}$:

\begin{equation}
\begin{split}
f(\bm{P}) &= -\sum_{i=1}^{n} \sum_{j=1}^n \langle \bm{A}_{ij},\bm{P}_{ij} \rangle + \lambda \cdot \mbox{rank}(\bm{P}), \\
&= -\langle \bm{A},\bm{P} \rangle + \lambda \cdot \mbox{rank}(\bm{P}),
\end{split}
\end{equation}
where $\bm{A}$ is concatenation of all $\bm{A}_{ij}$ similar to the form in \eqref{eq:big_p}, $\lambda$ denotes the weight of low-rank constraint.

The benefits of formulating the problem in this way are two-fold. First, the cycle consistency constraint aggregates the multi-way information to improve the matching and prune the false detections, which can hardly be realized if only two views are considered. Second, the rank minimization will automatically recover a rank (the number of people in the scene) that can best explain the observations.

\paragraph{Optimization:} To make the optimization tractable, we have to make appropriate relaxations. Instead of minimizing the rank, which is a discrete operator, we minimize the nuclear norm $\|\bm{P}\|_*$, which is the tightest convex surrogate of rank \cite{fazel2002matrix}. We replace the integer constraint on $\bm{P}$ by saying that $P$ is a real matrix with values in $[0,1]$:
\begin{align}
0 \leq \bm{P} \leq 1, \label{eq:value in zero to one}
\end{align}
which is a common practice in matching algorithms. We remove the semidefinite constraint and only require $\bm{P}$ to be symmetric:
\begin{align}
&\bm{P}_{ij} = \bm{P}_{ji}^T,~~1 \leq i,j \leq n, i \neq j, \label{eq:symmetric} \\
&\bm{P}_{ii} = \bm{I}_{p_i},~~~~1 \leq i \leq n. \label{eq:identity} 
\end{align}

Finally, we solve the following optimization problem:

\begin{equation}\label{eq:formulation}
\begin{split}
\min_{\bm{P}} \ -&\langle \bm{A},\bm{P} \rangle + \lambda \|\bm{P}\|_*, \\
\text{s.t.} ~~~~ \ &\bm{P} \in \mathcal C,
\end{split}
\end{equation}
where $\mathcal{C}$ denotes the set of matrices satisfying the constraints \eqref{eq:doubly stochastic constraints}, \eqref{eq:value in zero to one},  \eqref{eq:symmetric}, and \eqref{eq:identity}.

Note that the problem in \eqref{eq:formulation} is convex and we use the alternating direction method of multipliers (ADMM) \cite{boyd2011distributed} to solve it. The problem is first rewritten as follows by introducing an auxiliary variable $\bm{Q}$:
\begin{equation}\label{eq:rewrite}
\begin{split}
\min_{\bm{P},\bm{Q}} \ -&\langle \bm{A},\bm{P} \rangle + \lambda \|\bm{Q}\|_*, \\
\text{s.t.} ~~~~ \ &\bm{P} = \bm{Q}, \  \bm{P} \in \mathcal C.
\end{split}
\end{equation}

Then, the augmented Lagrangian of \eqref{eq:rewrite} is:
\begin{equation}\label{eq:augmented}
\begin{split}
\mathcal{L}_{\rho}(\bm{P},\bm{Q},\bm{Y}) &= -\langle \bm{A},\bm{P} \rangle + \lambda \|\bm{Q}\|_*  + \langle \bm{Y},\bm{P}-\bm{Q} \rangle \\ &+ \frac{\rho}{2}\|\bm{P}-\bm{Q}\|_{F}^{2},
\end{split}
\end{equation}
where $\bm{Y}$ denotes the dual variable and $\rho$ denotes a penalty parameter.
Each primal variable and the dual variable are alternately updated until convergence. The overall algorithm is shown in Algorithm \ref{alg:matchSVT}, where $\mathcal{D}$ denotes the operator for singular value thresholding \cite{cai2010singular}  

and $\mathcal{P}_{\mathcal{C}}(\cdot)$ denotes the orthogonal projection to $\mathcal{C}$.

\begin{algorithm}\caption{Consistent Multi-Way Matching }\label{alg:matchSVT}
	\LinesNumbered
	\KwIn{Affinity matrix $\bm{A}$}
	\KwOut{Consistent correspondences $\bm{P}$}
	randomly initialize $\bm{P}$ and $\bm{Y}=\bm{0}$ ;\\
	\While{not converged}{
		$\bm{Q} \leftarrow \mathcal{D}_{\frac{\lambda}{\rho}}(\frac{1}{\rho}\bm{Y}+\bm{P})$ ; \\
		$\bm{P} \leftarrow \mathcal{P}_{\mathcal{C}}(\bm{Q}-\frac{1}{\rho}(\bm{Y}-\bm{A}))$ ; \\
		$\bm{Y} \leftarrow \bm{Y}^k + \rho(\bm{P}-\bm{Q})$ ;
		
	}
	quantize $\bm{P}$ with a threshold equal to 0.5.
\end{algorithm}

The output $\bm{P}$ gives us the cycle-consistent correspondences of bounding boxes across all views. Figure \ref{fig:pipeline} shows an example. The bounding boxes with no matches in other views are regarded as false detections and discarded. 

%-------------------------------------------------------------------------

\subsection{3D pose reconstruction}\label{sec:multi-view}
Given the estimated 2D poses of the same person in different views, we reconstruct the 3D pose. This can be simply done by triangulation, but the gross errors in 2D pose estimation may largely degrade the reconstruction. In order to fully integrate uncertainties in 2D pose estimation and incorporate the structural prior on human skeletons, we make use of the 3DPS model and propose an approximate algorithm for efficient inference. 

\paragraph{3D pictorial structure:} We use a joint-based representation of 3D poses, i.e., $\bm{T} = \{\bm{t}_i | i=1,...,N\}$, where $\bm{t}_i \in \mathbb{R}^{3}$ denotes the location of joint $i$. Given 2D images from multiple views $I=\{I_v | v=1,...,V\}$, the posterior distribution of 3D poses can be written as:
\begin{equation}
p(\bm{T}|I) \propto \prod_{v=1}^{V} \prod_{i=1}^{N} p(I_v|\pi_v(\bm{t}_i))\prod_{(i,j) \in \varepsilon} p(\bm{t}_i,\bm{t}_j),
\end{equation}
where $\pi_v(\bm{t}_i)$ denotes the 2D projection of $\bm{t}_i$ in the $v$-th view and the likelihood $p(I_v|\pi_v(\bm{t}_i))$ is given by the 2D heat map output by the CNN-based 2D pose detector \cite{chen2017cascaded}, which characterizes the 2D spatial distribution of each joint.

The prior term $p(\bm{t}_i,\bm{t}_j)$ denotes the structural dependency between joint $\bm{t}_i$ and $\bm{t}_j$, which implicitly constrains the bone length between them. Here, we use a Guassian distribution to model the prior on bone length:
\begin{equation}
p(\bm{t}_i,\bm{t}_j) \propto N(\|\bm{t}_i-\bm{t}_j\| |L_{ij},\sigma_{ij}),
\end{equation}
where $\|\bm{t}_i-\bm{t}_j\|$ denotes the Euclidean distance between joint $\bm{t}_i$ and $\bm{t}_j$, $L_{ij}$ and $\sigma_{ij}$ denote the mean and standard deviation  respectively, learned from the Human3.6M dataset~\cite{ionescu2014human3}.

\paragraph{Inference:} The typical strategy to maximize $p(\bm{T}|I)$ is first discretizing the state space as a uniform 3D gird, and applying the max-product algorithm \cite{burenius20133d,pavlakos2017harvesting}. However, the complexity of the max-product algorithm grows fast with the dimension of the state space. 

Instead of using grid sampling, we set the state space for each 3D joint to be the 3D proposals triangulated from all pairs of corresponding 2D joints. As long as a joint is correctly detected in two views, its true 3D location is included in the proposals. In this way, the state space is largely reduced, resulting in much faster inference without sacrificing the accuracy.

%% file: 04_empirical_evaluation.tex
%!TEX root = ./egpaper_for_review.tex

We evaluate the proposed approach on three public datasets including both indoor and outdoor scenes and compare it with previous works as well as several variants of the proposed approach.

%-------------------------------------------------------------------------
\subsection{Datasets}

The following three datasets are used for evaluation:

\noindent \textbf{Campus}\cite{belagiannis20143d}: It is a dataset consisting of three people interacting with each other in an outdoor environment, captured with three calibrated cameras. We follow the same evaluation protocol as in previous works \cite{belagiannis20143d, belagiannis2014multiple, belagiannis20163d, ershadi2018multiple} and use the percentage of correctly estimated parts (PCP) to measure the accuracy of 3D location of the body parts.

\noindent \textbf{Shelf}\cite{belagiannis20143d}: Compared with Campus, this dataset is more complex, which consists of four people disassembling a shelf at a close range. There are five calibrated cameras around them, but each view suffers from heavy occlusion. The evaluation protocol is as the same as the prior work, and the evaluation metric is also 3D PCP.

\noindent \textbf{CMU Panoptic}\cite{joo2015panoptic}: This dataset is captured in a studio with hundreds of cameras, which contains multiple people engaging in social activities. For the lack of ground truth, we qualitatively evaluate our approach on the CMU Panoptic dataset.

%-------------------------------------------------------------------------
\subsection{Ablation analysis}
\iffalse
\begin{figure}[t]
	\centering
	\includegraphics[width=1.4\linewidth,trim={3cm 9cm 0.0cm 2cm},clip]{Figures/typical_case.pdf}
	\caption{
		A typical case of occlusion in the Campus dataset. The yellow bounding box represents the actor 3, who is only visible in two cameras.
	}
	\label{fig:typical_case}
\end{figure}
\fi
We first give an ablation analysis to justify the algorithm design in the proposed approach. The Campus and Shelf datasets are used for evaluation.

\paragraph{Appearance or geometry?} As described in section \ref{sec:multi-view people matching}, our approach combines appearance and geometry information to construct the affinity matrix. Here, we compare it with the alternatives using appearance or geometry alone. The detailed results are presented in Table \ref{tb:ablative}.

On the Campus, using appearance only achieves competitive results, since the appearance difference between actors is large. The result of using geometry only is worse because the cameras are far from the people, which degrades the discrimination ability of the epipolar constraint. On the Shelf, the performance of using appearance alone drops a lot. Especially, the result of actor 2 is erroneous, since his appearance is similar to another person. In this case, the combination of appearance and geometry greatly improve the performance.

\begin{table}[t]
	\begin{center}
		\setlength{\tabcolsep}{2pt}
		\begin{tabular}{rcccc}\cline{1-5}
			Campus & Actor 1 & Actor 2 & Actor 3 & Average  \\\hline
			Ours & {\bf 97.6} & {\bf 93.3} & {\bf 98.0} & {\bf 96.3 } \\
			Appearance  & {\bf 97.6} & {\bf 93.3} & 96.5 & 95.8 \\
			Geometry & 97.4 & 90.1 & 89.4 & 92.3 \\
			No 3DPS  & 90.6 & 89.2 & 97.7 & 92.5 \\
			No matching & 84.8 & 89.0 & 71.5 & 81.8 \\\hline
			Shelf & Actor 1 & Actor 2 & Actor 3 & Average  \\\hline
			Ours & {\bf 98.8} & {\bf 94.1} & {\bf 97.8} & {\bf 96.9 } \\
			Appearance  & 98.6 & 60.5 & 94.3 & 84.5 \\
			Geometry & 97.2 & 79.5 & 96.5 &  91.1\\
			No 3DPS & 97.9 & 89.5 & {\bf 97.8} & 95.1 \\
			No matching & 98.1 & 91.1 & 92.8 & 94.0 \\\hline
		\end{tabular}
	\end{center}
	\vspace{-5pt}
	\caption{Ablative study on the Campus and Shelf datasets. Appearance and geometry denote the different types of affinity matrices, i.e., using appearance only and using geometry only. `No 3DPS' uses triangulation instead of the 3DPS model to reconstruct 3D poses. `No matching' represents the 3DPS model without bounding box matching, an approach typically used in previous methods \cite{belagiannis20163d,joo2017panoptic}. We re-implement this approach with the state-of-the-art 2D pose detector. The numbers are the percentage of correctly estimated parts (PCP).}\label{tb:ablative}
\end{table}

\paragraph{Direct triangulation or 3DPS?} Given the matched 2D poses in all views, we use a 3DPS model to infer the final 3D poses, which is able to integrate the structural prior on human skeletons.
A simple alternative is to reconstruct 3D pose by triangulation, i.e., finding the 3D pose that has the minimum reprojection errors in all views.
The result of this baseline method (`NO 3DPS') is presented in Table \ref{tb:ablative}.

The result shows that when the number of cameras in the scene is relatively small, for example, in the Campus dataset (three cameras), using 3DPS can greatly improve the performance. When a person is often occluded in many views, for example, actor 2 in the Shelf dataset, the 3DPS model can also be helpful.

\paragraph{Matching or no matching?} Our approach first matches 2D poses across views and then applies the 3DPS model to each cluster of matched 2D poses. An alternative approach in most previous works \cite{belagiannis20163d,joo2017panoptic} is to directly apply the 3DPS model to infer multiple 3D poses from all detected 2D poses without matching. Here, we give a comparison between them. As Belagiannis \etal \cite{belagiannis20163d} did not use the most recent CNN-based keypoint detectors and Joo \etal \cite{joo2017panoptic} did not report results on public benchmarks, we re-implement their approach with the state-of-the-art 2D pose detector \cite{cao2016realtime} for a fair comparison. The implementation details are given in the supplementary materials. Table \ref{tb:ablative} shows that the 3DPS without matching obtained decent results on the Self dataset but performed much worse on the Campus dataset, where there are only three cameras. The main reason is that the 3DPS model implicitly uses multi-view geometry to link the 2D detections across views but ignores the appearance cues. When using a sparse set of camera views, the multi-view geometric consistency alone is sometimes insufficient to differentiate the correct and false correspondences, which leads to false 3D pose estimation. This observation coincides with the other results in Table \ref{tb:ablative} as well as the observation in \cite{joo2017panoptic}. The proposed approach explicitly leverage the appearance cues to find cross-view correspondences, leading to more robust results. Moreover, the matching step significantly reduces the size of state space and makes the 3DPS model inference much faster.

%-------------------------------------------------------------------------
\subsection{Comparison with state-of-the-art}\label{sec:comparison}

We compare with the following baseline methods.

Belagiannis \etal \cite{belagiannis20143d,belagiannis2014multiple} were among the first to introduce 3DPS model-based multi-person pose estimation and was extended to the video case to leverage temporal consistency \cite{belagiannis20163d}. Ershadi-Nasab \etal \cite{ershadi2018multiple} is a very recent method that proposes to cluster the 3D candidate joints to reduce the state space.

The results on the Campus and Shelf datasets are presented in Table \ref{tb:state-of-the-art}.
Note that the 2D pose detector \cite{chen2017cascaded} and the reID network \cite{zhong2018camera} used in our approach are the released pre-trianed models without any fine-tuning on the evaluated datasets. Even with the generic models, our approach outperforms the state-of-the-art methods by a large margin. In particular, our approach significantly improves the performance on the actor 3 in the Campus dataset and the actor 2 in the Shelf dataset, who suffer from severe occlusion. We also include our results without the 3DPS model but using triangulation to reconstruct 3D poses from matched 2D poses. Thanks to the robust and consistent matching, direct triangulation also obtains better performance than previous methods.

\begin{table}[t]
	\begin{center}
		\setlength{\tabcolsep}{2pt}
		\begin{tabular}{rcccc}\cline{1-5}
			Campus & Actor 1 & Actor 2 & Actor 3 & Average  \\\hline
			Belagiannis \etal \cite{belagiannis20143d}  & 82.0 & 72.4 & 73.7 & 75.8 \\
			Belagiannis \etal \cite{belagiannis2014multiple} & 83.0 & 73.0 & 78.0 & 78.0 \\
			Belagiannis \etal \cite{belagiannis20163d} & 93.5 & 75.7 & 84.4 & 84.5 \\
			Ershadi-Nasab \etal \cite{ershadi2018multiple} & 94.2 & 92.9 & 84.6 & 90.6 \\
			Ours w/o 3DPS  & 90.6 & 89.2 & 97.7 & 92.5 \\
			Ours  & {\bf 97.6} & {\bf 93.3} & {\bf 98.0} & {\bf 96.3 } \\\hline
			Shelf & Actor 1 & Actor 2 & Actor 3 & Average  \\\hline
			Belagiannis \etal \cite{belagiannis20143d}  & 66.1 & 65.0 & 83.2 & 71.4 \\
			Belagiannis \etal \cite{belagiannis2014multiple} & 75.0 & 67.0 & 86.0 & 76.0 \\
			Belagiannis \etal \cite{belagiannis20163d} & 75.3 & 69.7 & 87.6 & 77.5 \\
			Ershadi-Nasab \etal \cite{ershadi2018multiple} & 93.3 & 75.9 & 94.8 & 88.0 \\
			Ours w/o 3DPS & 97.9 & 89.5 & {\bf 97.8} & 95.1 \\
			Ours  & {\bf 98.8} & {\bf 94.1} & {\bf 97.8} & {\bf 96.9 }  \\\hline
			
		\end{tabular}
	\end{center}
	\vspace{-5pt}
	\caption{Quantitative comparison on the Campus and Shelf datasets. The numbers are percentage of correctly estimated parts (PCP). The results of other methods are taken from respective papers. `Ours w/o 3DPS' means using triangulation instead of the 3DPS model to reconstruct 3D poses from matched 2D poses.}
	\label{tb:state-of-the-art}
\end{table}

%-------------------------------------------------------------------------

\begin{figure*}[t]
	% minipage mit (Blind-)Text
	\centering
	\includegraphics[width=1.0\linewidth,trim={1.5cm 6.5cm 2cm 1cm},clip]{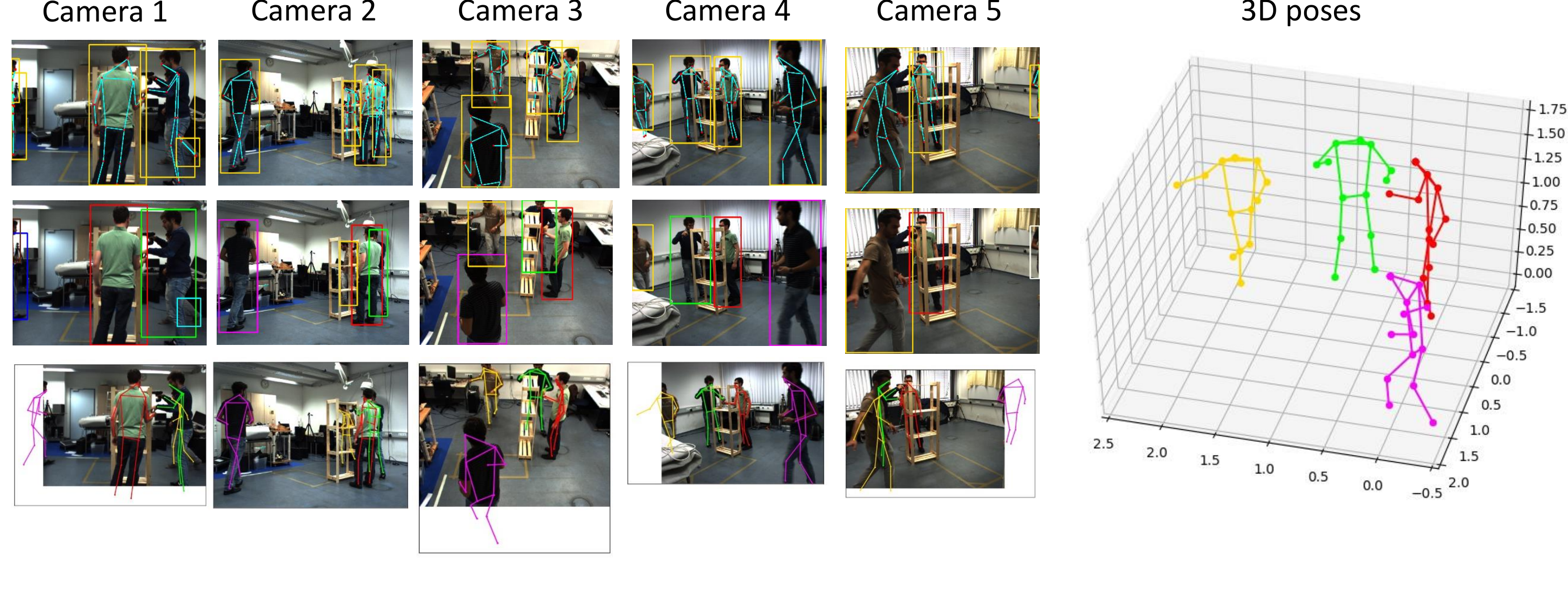}
	\includegraphics[width=1\linewidth,trim={1.5cm 5.8cm 2cm 1cm},clip]{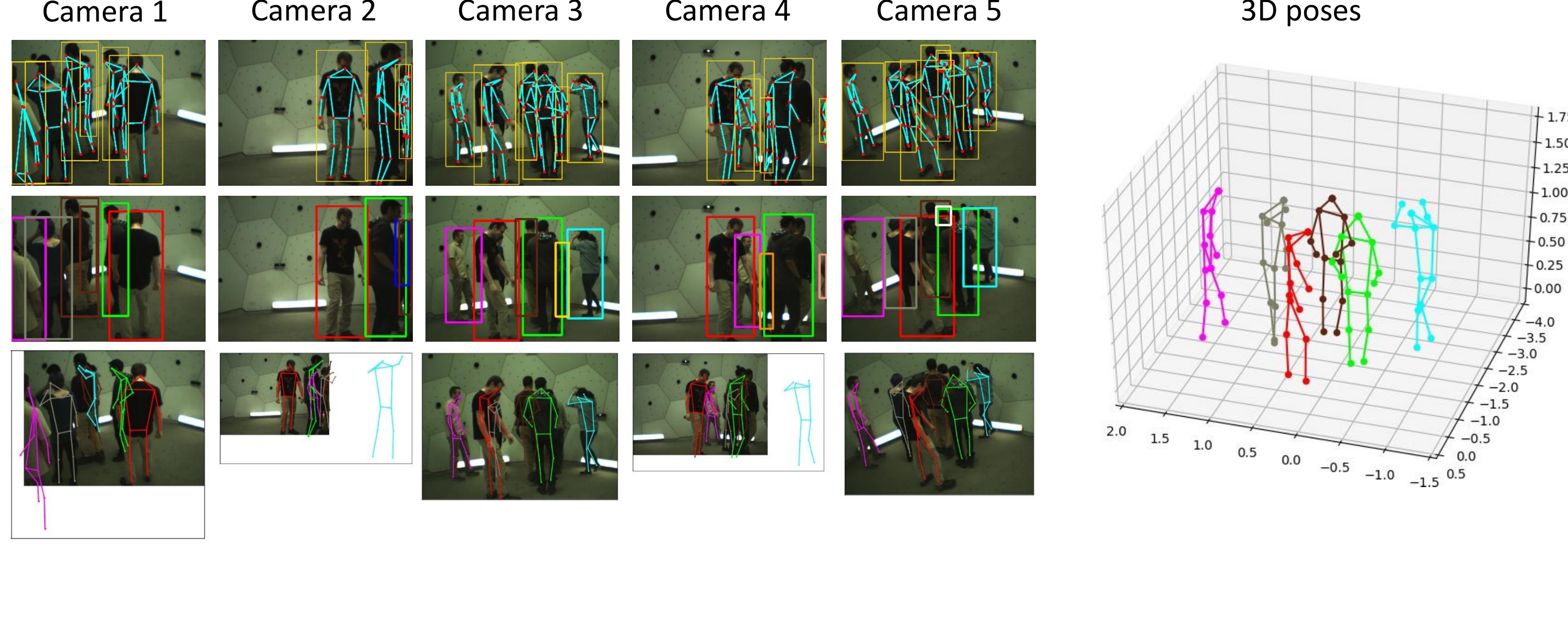}
	\vspace{-3em}
	\caption{
		Qualitative results on the Shelf (top) and CMU panoptic (bottom) datasets. The first row shows the 2D bounding box and pose detections. The second row shows the result of our matching algorithm where the colors indicate the correspondences of bounding boxes across views. The third row shows the 2D projections of the estimated 3D poses.
	}
	\label{fig:qual1}
\end{figure*}

\subsection{Qualitative evaluation}

Figure \ref{fig:qual1} shows some representative results of the proposed approach on the Shelf and CMU Panoptic dataset. Taking inaccurate 2D detections as input, our approach is able to establish their correspondences across views, identity the number of people in the scene automatically, and finally reconstruct their 3D poses. The final 2D pose estimates obtained by projecting the 3D poses back to 2D views are also much more accurate than the original detections.

\subsection{Running time}

We report running time of our algorithm on the sequences with four people and five views in the Shelf dataset, tested on a desktop with an Intel i7 3.60 GHz CPU and a GeForce 1080Ti GPU. Our unoptimized implementation on average takes 25 ms for running reID and constructing affinity matrices, 20 ms for the multi-way matching algorithm, and 60 ms for 3D pose inference. Moreover, the results in Table~\ref{tb:state-of-the-art} show that our approach without the 3DPS model also obtains very competitive performance, which is able to achieve real-time performance at $>20$fps.

%% file: 05_summary.tex
%!TEX root = ./egpaper_for_review.tex

In this paper, we propose a novel approach to multi-view 3D pose estimation that can fastly and robustly recover 3D poses of a crowd of people with a few cameras. Compared with the previous 3DPS based methods, our key idea is to use a multi-way matching algorithm to cluster the detected 2D poses to reduce the state space of the 3DPS model and thus improves both efficiency and robustness. We also demonstrate that the 3D poses can be reliably reconstructed from clustered 2D poses by triangulation even without using the 3DPS model. This shows the effectiveness of the proposed multi-way matching algorithm, which leverages the combination of geometric and appearance cues as well as the cycle-consistency constraint for matching 2D poses across multiple views.  